\documentclass[runningheads]{llncs}
\usepackage[T1]{fontenc}
\usepackage{graphicx}
\usepackage{marvosym}
\usepackage{xcolor}
\definecolor{freqgray}{gray}{0.25}
\usepackage{booktabs}    
\usepackage{tabularx}    
\usepackage{array}       

\usepackage{amsmath}
\usepackage{amssymb}
\usepackage{placeins}
\usepackage[skins, breakable]{tcolorbox} 
\usepackage{enumitem}                    
\usepackage{float}
\usepackage[shortcuts]{extdash}
\usepackage{arydshln}
\usepackage{multirow}
\usepackage{arydshln} 
\begin{document}

\title{ScienceJudge: Pattern Perturbation Augmentation for Sentence-Level Scientific Claim Verification}
\titlerunning{ScienceJudge}
%
\author{
Chuhai Cai\inst{1} \and
Xin Chen\inst{1} \and
Min Yang\inst{1,2} \and
Feng Jiang\inst{1}\textsuperscript{(\Letter)}
}
\authorrunning{C. Cai et al.}
%
\institute{
Institute of Artificial Intelligence, Shenzhen University of Advanced Technology, Shenzhen, China\\
\email{
\{24000217,25070145\}@stu.suat-sz.edu.cn, jiangfeng@suat-sz.edu.cn} \and
Shenzhen Institutes of Advanced Technology, Chinese Academy of Sciences, Shenzhen, China\\
\email{min.yang@siat.ac.cn}}







%
\maketitle
\begin{abstract}
Sentence-level scientific claim verification determines whether each sentence in an AI-generated scientific paragraph is supported by a given multimodal evidence bundle. Fine-grained reporting errors often preserve fluent and plausible statements while introducing subtle evidence--claim deviations, making them difficult to detect under sparse and imbalanced supervision. We propose \textbf{Pattern Perturbation Augmentation (PPA)}, which models reporting errors as label-specific faithfulness breaking transformations and induces reusable error patterns from labeled examples. PPA applies these patterns to supported claims to construct controlled hard negatives and expands their coverage through pattern-level, operation-level, and cross-domain augmentation. Based on these augmented views, we develop \textsc{ScienceJudge}, which integrates PPA-trained vision-language predictors and instruction-guided LLM predictors through class-weighted ensemble inference. On the local development set, \textsc{ScienceJudge} improves the average score from $0.8124$ to $0.8692$, achieving a Macro-F1 of $0.8466$ and a Paragraph Exact Match score of $0.8919$. Our submitted system ranks first on the official NLPCC 2026 Task~10 Track~1 leaderboard with a final score of $0.6890$.
\end{abstract}

\section{Introduction}

Large language models (LLMs) are increasingly used to assist scientific writing and report generation~\cite{birhane2023science,boyko2023interdisciplinary,wu-etal-2025-survey}.
However, the generated content may not faithfully reflect the underlying data and evidence.
Scientific reporting errors are often subtle: a generated statement may remain fluent and scientifically plausible while misattributing an entity, overstating the scope of a finding, reversing a reported relation, or introducing an unsupported causal interpretation.
Detecting such errors is therefore essential for reliable AI-assisted scientific writing.

Existing studies generally formulate this problem as \emph{evidence-based claim verification}, where models determine whether a claim is supported by the available evidence~\cite{thorne2018fever,wadden2020fact,vladika2023scientific}.
Prior work has explored rationale modeling, full-document context, multimodal verification over figures and tables, and hard-negative augmentation for robustness~\cite{wadden2022multivers,wang2025sciver,ho2026sciclaimeval,schuster2019towards,kaushik2020learning,lee2021crossaug}.
These approaches learn claim--evidence mappings for verification, but
their augmentation mechanisms rarely specify which evidence-grounded
element should change, how the claim--evidence relation should change,
or under which conditions an edit realizes a designated reporting-error
category. Consequently, broad perturbations can provide limited
coverage of the fine-grained error constructions needed for this task,
particularly when negative instances are scarce and imbalanced.

To address this gap, we propose \textbf{Pattern Perturbation
Augmentation (PPA)}, a structured framework that induces
label-specific faithfulness-breaking patterns from observed reporting
errors. PPA comprises three complementary strategies.
\emph{Pattern-Level Instantiation} retains a complete pattern---its
target slot, relation change, and applicability conditions---to realize
fine-grained error constructions. \emph{Operation-Level Abstraction}
derives reusable relation-changing operations from related patterns and
applies them across different targets and scientific contexts.
\emph{Cross-Domain Transfer} realizes these operations on evidence from
additional scientific domains. \emph{Scaled Operation PPA} is an
operation-scale variant that increases the accepted realizations under
the same operation inventory. Building on these complementary views, we
develop \textsc{ScienceJudge}, which combines separately trained vision-language predictors with instruction-guided LLM predictors through class-weighted ensemble inference.

We evaluate the proposed approach on Track~1 of the NLPCC 2026 Shared Task~10,\footnote{\url{https://nlp2ct.github.io/NLPCC-2026-Task10-Science}}
which requires systems to verify each sentence in an AI-generated scientific paragraph against a provided multimodal evidence bundle.
On our fixed local development set, \textsc{ScienceJudge} improves the average score from $0.8124$ under standard fine-tuning to $0.8692$, with a Macro-F1 of $0.8466$ and a Paragraph Exact Match score of $0.8919$.
The submitted system ranks first on the official Track~1 final leaderboard with a score of $0.6890$.
Further analysis shows that preserving fine-grained patterns sharpens category distinctions, cross-domain transfer improves scope recognition, and class-weighted voting prevents diagnostic unsupported predictions from being suppressed by the dominant \emph{Supported} label.

Our main contributions are summarized as follows:
\begin{itemize}
\item We analyze fine-grained scientific reporting errors and induce
label-specific faithfulness-breaking patterns that expose recurring
evidence deviations under scarce and imbalanced supervision.
\item We propose PPA, a structured hard-negative augmentation framework
that combines Pattern-Level Instantiation, Operation-Level Abstraction,
and Cross-Domain Transfer; an operation-scale variant
further examines additional contextual coverage.
\item We develop \textsc{ScienceJudge}, integrating PPA-trained vision-language predictors and instruction-guided LLM predictors through class-weighted ensemble inference, and achieve first place on NLPCC 2026 Shared Task~10 Track~1.
\end{itemize}

\section{Related Work}

\subsection{Evidence-Based Fact and Scientific Claim Verification}

Evidence-based fact verification determines claim veracity by grounding predictions in retrieved evidence~\cite{vlachos2014fact,guo2022survey}. FEVER established the influential retrieval--selection--verification paradigm~\cite{thorne2018fever}, while SciFact and MultiVerS extended verification to scientific literature with rationale supervision and full-document context~\cite{wadden2020fact,wadden2022multivers}. Recent studies further explore explainable verification over long-context documents and evidence-grounded scientific attribution, including FinDVer and SciTrue~\cite{zhao2024findver,tan2026scitrue}. Unlike prior settings that mainly formulate verification as claim-level veracity prediction, our task provides the evidence bundle and requires fine-grained diagnosis of sentence-level reporting errors.

\subsection{Verification over Structured and Multimodal Evidence}

Fact verification has increasingly incorporated heterogeneous evidence. FEVEROUS combines textual and tabular reasoning~\cite{aly2021feverous}, while recent multimodal verification studies investigate evidence integration across text, images, and tables~\cite{wang2025piecing}. In scientific domains, SciVer evaluates claim verification over text, tables, and charts~\cite{wang2025sciver}, and MuSciClaims introduces figure-grounded scientific claims with manually designed perturbations for diagnostic evaluation~\cite{lal2025musciclaims}. Recent work also studies evidence-cell alignment and atomic reasoning for scientific table verification ~\cite{ho2025table,zhang2025atomic}. SciClaimEval further constructs cross-modal examples from authentic scientific claims and modified supporting evidence~\cite{ho2026sciclaimeval}. Together, these studies highlight the importance of multimodal grounding and fine-grained reasoning in scientific verification.

\subsection{Hard-Negative and Counterfactual Data Augmentation}

Verification models may exploit annotation artifacts and label-correlated expressions without adequately modeling the claim--evidence relation~\cite{schuster2019towards}. Counterfactual data augmentation addresses such shortcuts through minimal label-changing edits~\cite{kaushik2020learning}, while CrossAug constructs contrastive claim--evidence pairs for evidence-sensitive prediction~\cite{lee2021crossaug}. Recent scientific benchmarks additionally construct challenging examples through controlled claim perturbations or modifications to supporting tables and figures~\cite{lal2025musciclaims,ho2026sciclaimeval}. However, these approaches primarily focus on constructing diagnostic evaluation datasets, whereas PPA learns fine-grained, label-specific perturbation patterns from observed reporting errors and uses them for training augmentation.

\begin{figure*}[t]
    \centering
    \includegraphics[width=1\textwidth]{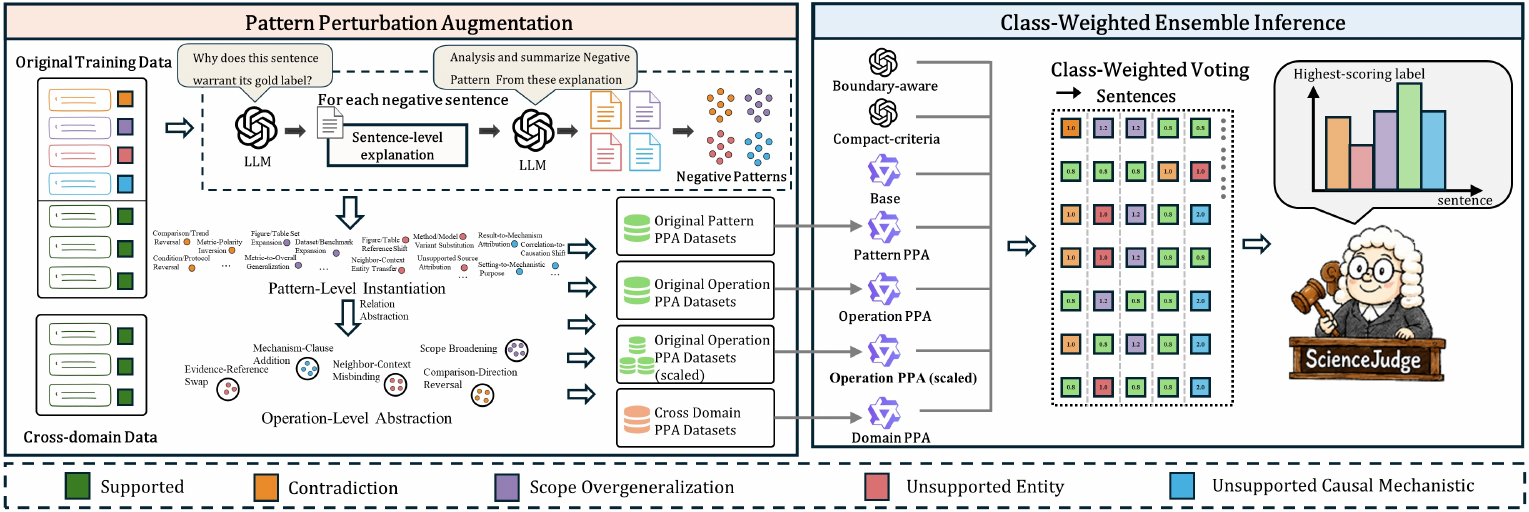}
    \caption{Overview of \textsc{ScienceJudge}. PPA induces label-specific
    error patterns from labeled training data and realizes them through
    Pattern-Level Instantiation, Operation-Level Abstraction, and
    Cross-Domain Transfer. Five local vision-language predictors, including
    a scaled operation variant, are combined with two instruction-guided
    LLM predictors through class-weighted ensemble inference.}
    \label{fig:methodology}
\end{figure*}

\section{Methodology}

\subsection{Task Formulation and Framework Overview}

Following the official formulation of NLPCC 2026 Shared Task~10
Track~1, the input consists of an AI-generated claim paragraph
$C=\{s_1,\ldots,s_n\}$ and an associated multimodal evidence bundle
$E$, which may contain textual descriptions, figure or table captions,
and associated images. The paragraph is segmented into $n$ sentences,
and the goal is to predict a sentence-level label sequence
$Y=\{y_1,\ldots,y_n\}$. Each $y_i$ is selected from five categories:
\emph{Supported}, \emph{Unsupported Entity}, \emph{Scope
Overgeneralization}, \emph{Contradiction}, and \emph{Unsupported
Causal Mechanistic}.

Figure~\ref{fig:methodology} presents \textsc{ScienceJudge}, an ensemble framework designed to address the challenges of fine-grained scientific reporting error identification. The key idea is to model reporting errors as recurring faithfulness-breaking transformations over evidence--claim relations and exploit them for controlled augmentation. Specifically, we propose \emph{Pattern Perturbation Augmentation} (PPA), which hierarchically explores these transformations by preserving fine-grained error structures, abstracting reusable relation-changing operations, and transferring them across scientific contexts. This design enables the model to capture label-specific error characteristics without being restricted to limited observed realizations. Finally, the augmented data are used to train complementary vision-language predictors, which are integrated with instruction-guided LLM predictors through class-weighted ensemble inference to improve robustness under highly imbalanced error categories.

\subsection{Negative Pattern Induction}
\label{sec:pattern-induction}

Unsupported scientific sentences often preserve most of a faithful
claim while changing one local relation between a claim and its
evidence. Through analysis of the training examples, we observe that
these deviations recur across error categories instead of appearing as
isolated generation errors. We therefore view them as label-specific
faithfulness-breaking transformations and represent each pattern as
\begin{equation}
p=\langle a,r,c\rangle,
\end{equation}
where $a$ is the affected semantic anchor, $r$ is the relation-changing
edit, and $c$ contains the applicability and label-specific decision
conditions. These conditions are natural-language rules that specify
suitable source sentences, permitted local edits, and distinctions among
neighboring labels. For example, a document-reference shift applies to
a Supported sentence with a concrete figure or table anchor; the edit
changes that anchor while preserving the proposition's scope, polarity,
and explanatory content.

To induce these patterns from observed errors, we analyze each negative
sentence in the official training data with GPT-5.5, providing the
evidence bundle, claim paragraph, target sentence, gold label, and
official label definition. The model identifies the altered semantic
element and explains how its relation to the evidence differs from that
of a faithful statement. A consolidation prompt groups the explanations
by their semantic anchors and relation-changing edits, while retaining
links to source examples. We then consolidate semantically equivalent
constructions and specify their applicability conditions against the
official definitions and examples. The resulting inventory includes
evidence-anchor substitution, figure--content misbinding, scope
expansion, polarity reversal, and unsupported mechanism insertion.

\subsection{Pattern Perturbation Augmentation}
\label{subsec:ppa}

Given the induced pattern inventory, PPA aims to alleviate supervision
sparsity by generating controlled hard negatives that preserve
label-specific evidence deviations while expanding their realizations.
It balances error specificity and transformation generalization:
Pattern-Level Instantiation maintains fine-grained error structures
within each label, whereas Operation-Level Abstraction derives reusable
relation-changing operations shared by related patterns. Cross-Domain
Transfer further extends these operations beyond the source scientific
distribution.

\subsubsection{In-Domain Data Augmentation}

Within the official scientific domain, PPA addresses two forms of
supervision sparsity: incomplete coverage of error structures within
each label and limited contextual variation of their underlying
evidence--claim deviations.

\textbf{Strategy I: Pattern-Level Instantiation.}
Fine-grained reporting labels often contain multiple
non-interchangeable evidence deviations. For \emph{Unsupported Entity},
these include shifted
evidence anchors, unsupported panel identifiers, figure--content
misbindings, and unsupported scientific objects. Although such cases
share a label, they rely on different local evidence cues.
Pattern-level instantiation retains the complete tuple
$\langle a,r,c\rangle$ as an augmentation unit. For each pattern,
compatible Supported sentences are selected using its applicability
conditions, and the prescribed local edit is instantiated. Sampling
across the pattern inventory distributes examples over frequent and
rare constructions. This path increases coverage of the distinct
error structures represented within each label.

\textbf{Strategy II: Operation-Level Abstraction.}
Different concrete patterns may share the same underlying
evidence--claim relation change while differing in their semantic
anchors and local contexts. We therefore abstract related patterns into
a reusable operation $o=g(p)$ that retains the label-changing relation
and its compatibility constraints. For example,
document-reference shifts and unsupported panel identifiers contribute
to evidence-anchor substitution; dataset-, setting-, and metric-level
expansions contribute to scope broadening; and result-to-mechanism
patterns contribute to unsupported mechanism insertion.
Operation-level augmentation applies an operation to compatible
Supported sentences while varying the semantic targets and local
scientific contexts. It therefore expands the contextual combinations
in which a stable evidence--claim violation is observed. Scaled
Operation PPA uses the same operation inventory, prompts, and quality
controls while accepting more realizations per operation; it is the
scale variant of this augmentation path.

\subsubsection{Cross-Domain Data Augmentation}

The official training data mainly cover computer science and AI
research. Cross-domain transfer is motivated by the observation that
evidence--claim violations can share structural patterns across domains
while their surface realizations vary across scientific fields. The
induced operations can therefore be transferred to additional domains
without domain-specific redesign.

We collect evidence-linked paragraphs from open-access articles in
biomedicine, materials science, and the physical sciences. Their
original evidence-grounded sentences provide Supported sources with
domain-specific terminology and reporting conventions. Given an induced
operation and a target label, PPA selects a compatible Supported
sentence from this collection and applies one local evidence--claim
transformation. The surrounding scientific content is preserved, so the
resulting negative realizes the same relation change in a different
domain context. A subset of the collected Supported sentences is
retained unchanged as controls.

\subsubsection{Controlled Instance Generation and Quality Control}

All PPA strategies share a controlled generation and filtering
procedure. Given an evidence bundle, a paragraph containing a compatible
\emph{Supported} sentence, a target label, and a transformation
specification, a language model generator produces one locally edited
sentence. The specification is a complete pattern for Pattern-Level
Instantiation and an abstract operation for the other two strategies.
Generation preserves the source topic, non-target scientific content,
and sentence fluency, and may abstain when no compatible target is
available.

We retain a candidate only when its source sentence can be uniquely
matched, exactly one sentence is modified, and the edit satisfies
length, sentence-boundary, and transformation-specific checks. These
checks distinguish nearby labels: changing a concrete evidence anchor
yields \emph{Unsupported Entity} when the proposition's applicability
is unchanged, whereas expanding that applicability yields \emph{Scope
Overgeneralization}. Duplicate outputs are removed, and the remaining
candidates are selected under strategy-specific quotas. We distinguish
candidates rejected because of inapplicability or failed checks from
valid candidates excluded only after the corresponding quota has been
reached.

\subsection{Instruction-Guided Predictors and Class-Weighted Ensemble Inference}
\label{sec:predictor-ensemble}

We complement the PPA-trained vision-language predictors with two
instruction-guided GPT-5.5 predictors used without fine-tuning. Both
receive the multimodal evidence, sentence-segmented claims, and label
definitions. The \emph{Boundary-aware} predictor distinguishes commonly
confused labels by locating the altered semantic element, whereas the
\emph{Compact-criteria} predictor applies concise evidence-grounded
criteria to each label.

\textsc{ScienceJudge} combines seven predictors: Base Training, four
PPA-trained vision-language predictors, and the two GPT-5.5 predictors.
Each produces one hard-label vote. For a sentence $x$, we predict
\begin{equation}
\hat{y}=\arg\max_{y\in\mathcal{Y}}
w_y\sum_{m\in\mathcal{M}}\mathbb{I}[f_m(x)=y],
\label{eq:ensemble}
\end{equation}
where $w_y$ is a class-specific vote multiplier. A predefined discrete
search on the development set selects
$(0.8,1.0,1.2,1.0,2.0)$ for Supported, Entity, Scope, Contradiction,
and Causal, respectively. The weights are fixed for test inference;
ties favor Contradiction, Causal, Entity, Scope, and Supported in order.

\section{Experiments}

\subsection{Experimental Settings}
\label{subsec:settings}

\subsubsection{Datasets and Metrics.}
As shown in Table~\ref{tab:data}, we use the Track~1 dataset of NLPCC 2026 Shared Task~10 on Scientific Claim Verification,\footnote{\url{https://github.com/NLP2CT/NLPCC-2026-Task10-Science}} which contains 3{,}333 paragraphs. We split the dataset into 3{,}000 paragraphs for training and 333 paragraphs for a fixed local development set using a stratified split that approximately preserves the original label distribution. The local development set contains 1{,}702 claim sentences and is used for model selection, prompt comparison, ensemble tuning, and local evaluation. We adopt the official sentence-level Macro-F1 and Paragraph Exact Match (PEM), with their average used as the overall score.

\begin{table}[t]
\centering
\caption{Statistics of the base training set and PPA datasets. Columns report sentence counts for each label and the number of paragraphs.}
\label{tab:data}
\small
\begin{tabular}{@{}lrrrrrr@{}}
\toprule
Data setting & Sup. & Entity & Scope & Contra. & Causal & Para. \\
\midrule
Base & 14{,}818 & 581 & 543 & 193 & 251 & 3{,}000 \\
Pattern PPA & 4{,}950 & 540 & 510 & 300 & 290 & 800 \\
Operation PPA & 2{,}113 & 200 & 100 & 100 & 100 & 500 \\
Operation PPA(scaled) & 10{,}849 & 1{,}000 & 500 & 500 & 500 & 2{,}500 \\
Domain PPA & 3{,}070 & 180 & 180 & 110 & 170 & 800 \\
\bottomrule
\vspace{-1cm}
\end{tabular}
\end{table}

\subsubsection{Training Corpora.}
Each PPA-based local vision-language predictor is trained on the official training data combined with the accepted augmentations from its corresponding PPA variant. We construct separate predictors for Pattern PPA, Operation PPA, Scaled Operation PPA, and Domain PPA. Scaled Operation PPA increases the number of accepted realizations per operation under the same generation procedure and quality constraints, while Merged PPA combines accepted augmentations from all variants.

\subsubsection{Implementation Details.}
All local vision-language predictors use Qwen3-VL-8B-Instruct and are fine-tuned with 4-bit QLoRA using LLaMA-Factory. LoRA uses rank 64 and alpha 128, with all eligible modules including visual components optimized during training.

We use a per-device batch size of 1 and gradient accumulation of 8.
Four-way distributed training on four NVIDIA RTX A6000 GPUs yields a
global batch size of 32. We use a learning rate of
$1.0\times10^{-4}$, a cosine learning-rate scheduler with a warmup
ratio of 0.1, bf16 precision, gradient checkpointing, and a maximum
image size of 262{,}144 pixels. We accessed GPT-5.5 through a commercial
API in June 2026 (\$43.38 development cost; no external retrieval at
test time).

\subsection{Main Results}
\label{subsec:main}

Table~\ref{tab:main} reports the development-set results of the individual predictors and ensemble systems.
\begin{table*}[!ht]
\centering
\caption{Development-set results of individual predictors and ensemble systems, including label-wise F1 scores. The \textbf{bold} indicates the best performance.}
\label{tab:main}
\renewcommand{\arraystretch}{1.15}
\resizebox{0.98\linewidth}{!}{%
\begin{tabular}{lcccccccc}
\toprule
\multirow{2}{*}{\textbf{Predictor$\downarrow$ Metric$\rightarrow$}}
& \multicolumn{5}{c }{\textbf{Label-wise F1}}
& \multicolumn{3}{c}{\textbf{Overall}} \\
\cmidrule(lr){2-6} \cmidrule(lr){7-9}
& \textbf{Sup.}
& \textbf{Entity}
& \textbf{Scope}
& \textbf{Contra.}
& \textbf{Causal}
& \textbf{Macro-F1}
& \textbf{PEM}
& \textbf{Avg.} \\
\midrule
\multicolumn{9}{c}{\textit{Text-only Surrogate}} \\
\cmidrule(lr){1-9}
Qwen3-8B-Instruct
& 0.7963 & 0.1754 & 0.0339 & 0.0696 & 0.1096 
& 0.2370 & 0.2523 & 0.2446
\\
+ Base Training
& 0.9784 & 0.7586 & 0.7833 & 0.5366 & 0.7692
& 0.7652 & 0.8078 & 0.7865
\\
\midrule
\multicolumn{9}{c}{\textit{Our PPA-Based Individual Model}} \\
\cmidrule(lr){1-9}
Qwen3-VL-8B-Instruct
& 0.8098 & 0.2872 & 0.0000 & 0.1154 & 0.0909
& 0.2606 & 0.2643 & 0.2625 \\
+ Base Training
& 0.9828 & 0.7290 & 0.8033 & 0.5778 & 0.8421
& 0.7870 & 0.8378 & 0.8124 \\
\quad + Pattern PPA
& 0.9847 & 0.8000 & 0.8000 & 0.6667 & 0.8254
& 0.8154 & 0.8649 & 0.8401 \\
\quad + Operation PPA
& 0.9831 & 0.7222 & 0.8099 & 0.6316 & 0.8197
& 0.7933 & 0.8438 & 0.8186 \\
\quad + Operation PPA(scaled)
& 0.9801 & 0.7647 & 0.7931 & 0.5385 & 0.8657
& 0.7884 & 0.8468 & 0.8176 \\
\quad + Domain PPA
& 0.9831 & 0.7966 & 0.8305 & 0.4348 & 0.8060
& 0.7702 & 0.8619 & 0.8160 \\
\quad \quad  Merged PPA
& 0.9824 & 0.7667 & 0.8125 & \textbf{0.7059} & 0.7419
& 0.8019 & 0.8529 & 0.8274 \\
\midrule
\multicolumn{9}{c}{\textit{Our Instruction-Guided Individual Model}} \\
\cmidrule(lr){1-9}
GPT-5.5 (Base)
& 0.7555 & 0.3180 & 0.2353 & 0.3467 & 0.1641
& 0.3639 & 0.1862 & 0.2751 \\
\quad + Boundary-aware Instruction
& 0.9614 & 0.6946 & 0.5909 & 0.5333 & \textbf{0.9000}
& 0.7360 & 0.7357 & 0.7359 \\
\quad + Compact-criteria Instruction
& 0.9446 & 0.7294 & 0.5424 & 0.5000 & 0.3061
& 0.6045 & 0.6396 & 0.6221 \\
\midrule
\multicolumn{9}{c}{\textit{Our Ensemble Model}} \\
\cmidrule(lr){1-9}
Local VLM Ensemble (5)
& 0.9860 & 0.8174 & 0.8305 & 0.6875 & 0.8387
& 0.8320 & 0.8799 & 0.8560 \\
Unweighted Voting (7)
& \textbf{0.9873} & \textbf{0.8430} & 0.8348 & 0.6875 & 0.8525
& 0.8410 & 0.8859 & 0.8634 \\
Class-Weighted Voting (7)
& \textbf{0.9873} & \textbf{0.8430} & \textbf{0.8496} & 0.6875 & 0.8657
& \textbf{0.8466} & \textbf{0.8919} & \textbf{0.8692} \\
\bottomrule
\end{tabular}%
}
\end{table*}

On the local development set, \textsc{ScienceJudge} reaches an average
score of \textbf{$0.8692$} (Macro-F1 $0.8466$; PEM $0.8919$), improving by
$0.0568$ over Base Training. The text-only Qwen3-8B surrogate trails
Qwen3-VL-8B both before task adaptation ($0.2446$ vs.\ $0.2625$) and
after Base Training ($0.7865$ vs.\ $0.8124$), consistent with
multimodal evidence providing useful verification cues. Fine-tuning raises the local VLM from $0.2625$ to $0.8124$, highlighting the importance of task-specific adaptation for learning the label distinctions and evidence--claim relations required by this task.

Beyond Base Training, Pattern PPA provides the strongest individual
local predictor, improving the average score from $0.8124$ to $0.8401$.
Its Entity and Contradiction F1 scores increase from $0.7290$ to
$0.8000$ and from $0.5778$ to $0.6667$, respectively. This result suggests that recurring evidence-claim deviation patterns can provide useful supervision beyond standard fine-tuning. Within the GPT-5.5 family, the
task-specific Boundary-aware and Compact-criteria prompts improve over
the generic prompt from $0.2751$ to $0.7359$ and $0.6221$,
respectively. The two instructions also provide different label
specialties: Boundary-aware achieves the strongest Causal F1 ($0.9000$),
whereas Compact-criteria attains higher Entity F1 ($0.7294$ vs.\
$0.6946$).

At the system level, the local VLM ensemble reaches $0.8560$,
unweighted voting over all predictors reaches $0.8634$, and
class-weighted voting reaches $0.8692$. These results suggest that combining PPA-trained vision-language predictors with instruction-guided LLM predictions is beneficial on the development set, while class weighting provides a further modest improvement. 

Table~\ref{tab:official-track1} reports the official final Track~1
leaderboard.\footnote{\url{https://nlp2ct.github.io/NLPCC-2026-Task10-Science/leaderboard.html}}
\textsc{ScienceJudge}, submitted as \textsc{SUAT-NLP}, ranks first with
an official final score of $0.6890$.
\begin{table}[t]
\centering
\caption{Top three systems on the official Track~1 leaderboard. The \textbf{Bold} indicates the best performance and \underline{underline} indicates the second best.
}
\label{tab:official-track1}
\scriptsize
\renewcommand{\arraystretch}{1.2}
\setlength{\tabcolsep}{2pt}
\begin{tabular}{@{}clccccccc@{}}
\toprule
\multirow{2}{*}{\textbf{Rank}} & \multirow{2}{*}{\textbf{System}} &
\multicolumn{3}{c}{\textbf{Phase 1}} &
\multicolumn{3}{c}{\textbf{Phase 2}} &
\multirow{2}{*}{\textbf{Avg.}} \\
\cmidrule(lr){3-5} \cmidrule(lr){6-8}
& & \textbf{Macro-F1} & \textbf{PEM} & \textbf{Avg.} &
\textbf{Macro-F1} & \textbf{PEM} & \textbf{Avg.} & \\
\midrule
\textbf{1} & \textbf{ScienceJudge (Ours)} &
\underline{0.5659} & \textbf{0.4710} & \textbf{0.5185} & \underline{0.9053} & \underline{0.8136} & \underline{0.8595} &
\textbf{0.6890} \\
2 & System 1 &
\textbf{0.5754} & \underline{0.4369} & \underline{0.5061} & \textbf{0.9056} & \textbf{0.8337} & \textbf{0.8697} &
\underline{0.6879} \\
3 & System 2 &
0.4996 & 0.3823 & 0.4409 & 0.7523 & 0.6854 & 0.7188 &
0.5799 \\
\bottomrule
\end{tabular}
\end{table}

\subsection{How Does PPA Change Local Error Recognition?}
\label{subsec:ppa-analysis}

We examine the prediction changes that reveal what the augmented
corpora add beyond aggregate scores.

\textbf{Pattern PPA makes reference identity a verifiable cue.}
Base Training predicts five \emph{Unsupported Entity} cases as
\emph{Supported} because their scientific propositions remain
compatible with the evidence topic; the error lies only in the identity
of the cited evidence object. Pattern PPA corrects all five decisions
by treating the figure or table reference anchor as part of the claim
semantics. The recovered cases include Figure~5 against evidence for
Figure~4, Table~5 against evidence for Table~4, and Figure~7 against
evidence for Figure~8. Its construction keeps the proposition and
local context fixed while replacing the reference anchor, teaching the
model that a scientifically plausible conclusion still forms an
\emph{Unsupported Entity} error when it is attributed to the wrong
evidence object.

This effect also reveals the broader mechanism of Pattern PPA. It
instantiates induced construction patterns across diverse scientific
statements: research content, semantic slots, and surface wording vary,
while the core evidence--claim relation that defines the label remains
unchanged. These contrasts expand the coverage of evidence--claim
configurations associated with each error family, enabling the model to
recognize the construction itself instead of relying on a particular
figure number, dataset name, or lexical form. The Figure/Table
corrections on the development set therefore provide an interpretable
window into the model's broader ability to recognize this family of
evidence deviations.

The same contrastive construction covers all labels: it tests
evidence alignment for \emph{Supported}, object identity for
\emph{Unsupported Entity}, mutually incompatible replacements for
\emph{Contradiction}, stated coverage for \emph{Scope
Overgeneralization}, and added mechanisms for \emph{Unsupported Causal
Mechanistic}. The visible Entity correction is one manifestation of broader pattern-recognition capability.

\textbf{Separate PPA views provide complementary augmentation perspectives.} Pattern PPA is designed to emphasize exact evidence binding, Operation PPA transfers relation changes across contexts, and Domain PPA exposes these transformations to different scientific language and reporting settings. These strategy-specific views are retained for subsequent decision-level aggregation.

\subsection{How Does Class-Weighted Ensemble Inference Affect Predictions?}
\label{subsec:ensemble-analysis}

\textbf{Decision-level fusion preserves complementary recognition
views.} Merged PPA reaches $0.8274$, showing that a pooled corpus
retains useful PPA signal, but it remains below the local VLM ensemble
($0.8560$). A merged predictor must encode all augmentation criteria in
one parameterization, whereas separate PPA-trained vision-language predictors preserve
distinct decisions for aggregation. Adding the two GPT-5.5 views further
raises the score to $0.8634$ despite their weaker standalone results:
they recover four missed \emph{Unsupported Entity} cases and return
three local false Scope or Causal alarms to \emph{Supported}. Their
contribution therefore comes from nonredundant evidence--claim
decisions rather than standalone strength.

\textbf{Class weighting acts on a small, identifiable ambiguity set.}
It changes six sentence labels relative to unweighted voting. Three
changes correct genuine label-specific deviations: a claim attributes a
larger low-loss region to reshaping the underlying optimization
geometry, a local ablation result is presented as representative of all
experiments, and a comparison combines an incorrect evidence anchor
with an unsupported explanation of an accuracy gap. One additional
change switches between two accepted labels. The two new errors instead
occur on a malformed one-token fragment and on a factual architectural
description that contains no causal claim. Thus, class weighting promotes
minority negative-label votes when they identify an explicit causal,
scope, or entity cue, while remaining vulnerable when sentence context
does not support a well-formed relation decision.

The three corrections complete two additional paragraphs, whereas
neither regression breaks an otherwise exact paragraph; this asymmetry
accounts for the PEM improvement from $0.8859$ to $0.8919$. The effect
is a narrow class-specific vote adjustment concentrated on ambiguous
evidence--claim relations.

\subsection{Augmentation Quality and Statistical Stability}
\label{subsec:quality-stability}

\textbf{Augmentation quality.}
We conduct a stratified manual audit of 100 retained augmentations,
with 25 examples sampled from each PPA strategy. Overall, 94\% of the
audited examples are clearly valid, preserving the intended
scientific content while introducing only the designated error type.
Pattern, Operation, and Scaled Operation PPA each achieve a 96\%
clearly-valid rate, while Domain PPA achieves 88\%. Most of the
remaining ambiguous cases arise from localized scope broadening,
suggesting that scope expansion is the primary residual source of
content drift. Prompts, augmentation inventories, filtering rules,
ensemble configurations, aggregate statistics, and analysis code are
publicly available.\footnote{\url{https://github.com/SUAT-AIRI/Science-Judge}}

\textbf{Statistical stability.}
We use 10{,}000 paragraph-level paired-bootstrap resamples and 10{,}000 paired-randomization trials to assess the robustness of the complete system. \textsc{ScienceJudge} improves over Base Training by $0.05684$ (95\% CI $[0.02428,0.09042]$; $p=0.0010$), with the confidence interval remaining entirely above zero. This result indicates that the overall improvement is stable under paragraph-level resampling and is unlikely to be explained by sampling variation alone. Because the same development split was used for system selection, we interpret this analysis as post-hoc evidence of statistical stability rather than independent confirmatory testing.

\section{Conclusion}

We presented \textsc{ScienceJudge}, a fine-grained scientific claim verification system built on \textbf{Pattern Perturbation Augmentation (PPA)} and decision-level ensemble inference. PPA constructs controlled hard negatives from recurring evidence--claim deviations, while \textsc{ScienceJudge} integrates complementary vision-language and instruction-guided LLM predictors. The system improves the local development score from $0.8124$ to $0.8692$ and ranks first on the official NLPCC 2026 Shared Task~10 Track~1 leaderboard with a final score of $0.6890$. Overall, our results suggest that structured, error-aware augmentation is a promising approach to scientific claim verification under sparse and imbalanced supervision.

\section{Limitations}
\label{sec:limitations}

The local development set is reused for model selection, ensemble tuning, and evaluation, so local comparisons and statistical tests are post-hoc rather than independent confirmatory evidence. Augmentation quality is evaluated on a manual audit of 100 retained examples, leaving room for larger-scale multi-annotator validation. Finally, GPT-5.5 is used for PPA pattern induction and as two instruction-guided ensemble predictors, and its transferability to alternative models and broader scientific domains remains to be studied.

%
%
%
\raggedbottom
\bibliographystyle{splncs04}
\bibliography{mybibliography}

@inproceedings{thorne2018fever,
  title = {{FEVER}: A Large-scale Dataset for Fact Extraction and Verification},
  author = {Thorne, James and Vlachos, Andreas and Christodoulopoulos, Christos and Mittal, Arpit},
  booktitle = {NAACL-HLT 2018},
  pages = {809--819},
  year = {2018},
}

@inproceedings{vlachos2014fact,
  title = {Fact Checking: Task Definition and Dataset Construction},
  author = {Vlachos, Andreas and Riedel, Sebastian},
  booktitle = {ACL 2014 Workshop on Language Technologies and Computational Social Science},
  pages = {18--22},
  year = {2014},
}

@inproceedings{schuster2019towards,
  title = {Towards Debiasing Fact Verification Models},
  author = {Schuster, Tal and Shah, Darsh and Yeo, Yun Jie Serene and Roberto Filizzola Ortiz, Daniel and Santus, Enrico and Barzilay, Regina},
  booktitle = {EMNLP-IJCNLP 2019},
  pages = {3419--3425},
  year = {2019},
}

@inproceedings{wadden2020fact,
  title = {Fact or Fiction: Verifying Scientific Claims},
  author = {Wadden, David and Lin, Shanchuan and Lo, Kyle and Wang, Lucy Lu and van Zuylen, Madeleine and Cohan, Arman and Hajishirzi, Hannaneh},
  booktitle = {EMNLP 2020},
  pages = {7534--7550},
  year = {2020},
}

@inproceedings{kaushik2020learning,
  title = {Learning the Difference that Makes a Difference with Counterfactually-Augmented Data},
  author = {Kaushik, Divyansh and Hovy, Eduard and Lipton, Zachary C.},
  booktitle = {ICLR 2020},
  year = {2020},
}

@article{birhane2023science,
  title = {Science in the Age of Large Language Models},
  author = {Birhane, Abeba and Kasirzadeh, Atoosa and Leslie, David and Wachter, Sandra},
  journal = {Nature Reviews Physics},
  volume = {5},
  number = {5},
  pages = {277--280},
  year = {2023},
}

@misc{boyko2023interdisciplinary,
  title = {An Interdisciplinary Outlook on Large Language Models for Scientific Research},
  author = {Boyko, James and Cohen, Joseph and Fox, Nathan and Veiga, Maria Han and Li, Jennifer I-Hsiu and Liu, Jing and Modenesi, Bernardo and Rauch, Andreas H. and Reid, Kenneth N. and Tribedi, Soumi and Visheratina, Anastasia and Xie, Xin},
  year = {2023},
  eprint = {2311.04929},
  archivePrefix = {arXiv},
  primaryClass = {cs.CL},
}

@inproceedings{aly2021feverous,
  title = {{FEVEROUS}: Fact Extraction and {VER}ification Over Unstructured and Structured Information},
  author = {Aly, Rami and Guo, Zhijiang and Schlichtkrull, Michael and Thorne, James and Vlachos, Andreas and Christodoulopoulos, Christos and Cocarascu, Oana and Mittal, Arpit},
  booktitle = {NeurIPS Datasets and Benchmarks 2021},
  year = {2021},
  eprint = {2106.05707},
  archivePrefix = {arXiv},
  primaryClass = {cs.CL},
}

@inproceedings{lee2021crossaug,
  title = {{CrossAug}: A Contrastive Data Augmentation Method for Debiasing Fact Verification Models},
author = {Lee, Minwoo and Won, Seungpil and Kim, Juae and Lee, Hwanhee
          and Park, Cheoneum and Jung, Kyomin},
  booktitle = {CIKM 2021},
  pages = {3181--3185},
  year = {2021},
}

@article{guo2022survey,
  title = {A Survey on Automated Fact-Checking},
  author = {Guo, Zhijiang and Schlichtkrull, Michael and Vlachos, Andreas},
  journal = {Transactions of the Association for Computational Linguistics},
  volume = {10},
  pages = {178--206},
  year = {2022},
}

@inproceedings{wadden2022multivers,
  title = {{MultiVerS}: Improving Scientific Claim Verification with Weak Supervision and Full-Document Context},
  author = {Wadden, David and Lo, Kyle and Wang, Lucy Lu and Cohan, Arman and Beltagy, Iz and Hajishirzi, Hannaneh},
  booktitle = {Findings of NAACL 2022},
  pages = {61--76},
  year = {2022},
}

@inproceedings{vladika2023scientific,
  title = {Scientific Fact-Checking: A Survey of Resources and Approaches},
  author = {Vladika, Juraj and Matthes, Florian},
  booktitle = {Findings of ACL 2023},
  pages = {6215--6230},
  year = {2023},
}

@inproceedings{wang2025sciver,
  title = {{SciVer}: Evaluating Foundation Models for Multimodal Scientific Claim Verification},
  author = {Wang, Chengye and Shen, Yifei and Kuang, Zexi and Cohan, Arman and Zhao, Yilun},
  booktitle = {ACL 2025},
  pages = {8562--8579},
  year = {2025},
}

@inproceedings{ho2026sciclaimeval,
  title = {{SciClaimEval}: Cross-modal Claim Verification in Scientific Papers},
  author = {Ho, Xanh and Wu, Yun-Ang and Kumar, Sunisth and Xia, Tian Cheng and Boudin, Florian and Greiner-Petter, Andr{\'e} and Aizawa, Akiko},
  booktitle = {LREC 2026},
  pages = {11060--11071},
  year = {2026},
}

@inproceedings{zhao2024findver,
  title = {{FinDVer}: Explainable Claim Verification over Long and Hybrid-content Financial Documents},
  author = {Zhao, Yilun and Long, Yitao and Jiang, Tintin and Wang, Chengye and Chen, Weiyuan and Liu, Hongjun and Tang, Xiangru and Zhang, Yiming and Zhao, Chen and Cohan, Arman},
  booktitle = {EMNLP 2024},
  pages = {14739--14752},
  year = {2024},
}

@inproceedings{lal2025musciclaims,
  title = {{MuSciClaims}: Multimodal Scientific Claim Verification},
  author = {Lal, Yash Kumar and Bandham, Manikanta and Hasan, Mohammad Saqib and Kashi, Apoorva and Koupaee, Mahnaz and Balasubramanian, Niranjan},
  booktitle = {IJCNLP-AACL 2025},
  pages = {3285--3307},
  year = {2025},
}

@inproceedings{ho2025table,
  title = {Table-Text Alignment: Explaining Claim Verification Against Tables in Scientific Papers},
  author = {Ho, Xanh and Kumar, Sunisth and Wu, Yun-Ang and Boudin, Florian and Takasu, Atsuhiro and Aizawa, Akiko},
  booktitle = {Findings of EMNLP 2025},
  pages = {2509--2517},
  year = {2025},
}

@misc{zhang2025atomic,
  title = {Atomic Reasoning for Scientific Table Claim Verification},
  author = {Zhang, Yuji and Wang, Qingyun and Qian, Cheng and Liu, Jiateng and Sun, Chenkai and Zhang, Denghui and Abdelzaher, Tarek and Zhai, Chengxiang and Nakov, Preslav and Ji, Heng},
  year = {2025},
  eprint = {2506.06972},
  archivePrefix = {arXiv},
  primaryClass = {cs.CL},
}

@inproceedings{wang2025piecing,
  title = {Piecing It All Together: Verifying Multi-hop Multimodal Claims},
  author = {Wang, Haoran and Rangapur, Aman and Xu, Xiongxiao and Liang, Yueqing and Gharwi, Haroon and Yang, Carl and Shu, Kai},
  booktitle = {COLING 2025},
  pages = {7453--7469},
  year = {2025},
}

@inproceedings{tan2026scitrue,
  title = {{SciTrue}: Evidence-Grounded Claim Verification in Science},
  author = {Tan, Ne{\c{s}}et and Li, Minghao and Gahegan, Mark},
  booktitle = {EACL 2026 (System Demonstrations)},
  pages = {397--406},
  year = {2026},
}

@article{wu-etal-2025-survey,
  title = {A Survey on {LLM}-Generated Text Detection: Necessity, Methods, and Future Directions},
  author = {Wu, Junchao and Yang, Shu and Zhan, Runzhe and Yuan, Yulin and Chao, Lidia Sam and Wong, Derek Fai},
  journal = {Computational Linguistics},
  volume = {51},
  number = {1},
  pages = {275--338},
  year = {2025},
}


\end{document}